\documentclass{article} 
\usepackage{nips13submit_e,times}
\usepackage{hyperref}
\usepackage{url}
\usepackage{amsmath,amssymb,amsfonts}
\usepackage{bm}

\DeclareMathOperator*{\argmin}{argmin}
\DeclareMathOperator*{\argmax}{argmax}

\def\beq{\begin{equation}}
\def\eeq{\end{equation}}

\title{Hybrid SRL with Optimization Modulo Theories}

\author{
Stefano Teso \quad\quad Roberto Sebastiani \quad\quad Andrea Passerini \\
Department of Information Engineering and Computer Science\\
University of Trento\\
Povo, Trento I-38123, Italy\\
\texttt{\{teso,roberto.sebastiani,passerini\}@disi.unitn.it}\\
}

\newcommand{\LAQ}{\ensuremath{\mathcal{LA}(\mathbb{Q})}}
\newcommand{\LAZ}{\ensuremath{\mathcal{LA}(\mathbb{Z})}}
\newcommand{\BV}{\ensuremath{\mathcal{BV}}}
\newcommand{\ST}{\ensuremath{\mathcal{ST}}}

\nipsfinalcopy 

\begin{document}

\maketitle

\begin{abstract}
Generally speaking, the goal of constructive learning could be seen as, given
an example set of structured objects, to generate {\em novel} objects with
similar properties. From a statistical-relational learning (SRL) viewpoint, the
task can be interpreted as a constraint satisfaction problem, i.e. the
generated objects must obey a set of soft constraints, whose weights are
estimated from the data. Traditional SRL approaches rely on (finite)
First-Order Logic (FOL) as a description language, and on MAX-SAT solvers to
perform inference.  Alas, FOL is unsuited for constructive problems where the
objects contain a mixture of Boolean and numerical variables. It is in fact
difficult to implement, e.g. linear arithmetic constraints within the language
of FOL. In this paper we propose a novel class of hybrid SRL methods that rely
on Satisfiability Modulo Theories, an alternative class of formal languages
that allow to describe, and reason over, mixed Boolean-numerical objects and
constraints. The resulting methods, which we call {\em Learning Modulo
Theories}, are formulated within the structured output SVM framework, and
employ a weighted SMT solver as an optimization oracle to perform efficient
inference and discriminative max margin weight learning. We also present a few
examples of constructive learning applications enabled by our method.
\end{abstract}

\section{Introduction}

Traditional statistical-relational learning (SRL) methods allow to reason and
make inference about relational objects characterized by a set of soft
constraints~\cite{getoor2007introduction}. Most methods rely on some form of
(finite) First-Order Logic (FOL) to encode the learning problem, and define the
constraints as weighted logical formulae. In this context, maximum a posteriori
inference is often interpreted as a (partial weighted) MAX-SAT problem, i.e.
finding a truth value {\em assignment} of all predicates that maximizes the
total weight of the satisfied formulae; moreover, MAX-SAT plays a role in
maximum likelihood inference as well. In order to solve this problem, SRL
methods may rely on one of the many efficient, approximate solvers available.
One issue with these approaches is that First-Order Logic is not suited for
reasoning over hybrid variables.  The propositionalization of an $n$-bit
integer variable requires $n$ distinct binary predicates, which account for
$2^n$ distinct states, making na\"ive translation impractical. In addition, FOL
offers no efficient mechanism to describe simple operators between numerical
variables, like comparisons (e.g.  ``less-than'', ``equal'') and arithmetical
operations (e.g.  summation), limiting the range of realistically applicable
constraints to those based solely on logical connectives.

In order to side-step these limitations, researchers in automated reasoning and
formal verification have developed more appropriate logical languages that
allow to {\em natively} reason over mixtures of Boolean {\em and} numerical
variables (or more complex algebraic structures). These languages are grouped
under the umbrella term of {\em Satisfiability Modulo Theories}
(SMT)~\cite{barrett2009satisfiability}.  Each such language corresponds to a
decidable fragment of First-Order Logic augmented with an additional background
theory $\mathcal{T}$. There are many such background theories, including those
of linear arithmetic over the rationals \LAQ\ and integers \LAZ, among
others~\cite{barrett2009satisfiability}. In SMT, a formula can contain Boolean
variables (i.e. logical predicates) and connectives, mixed with symbols defined
by the theory $\mathcal{T}$, e.g. rational variables and arithmetical
operators. For instance, the SMT(\LAQ) syntax allows to write constraints such
as:
$$ {\tt HasProperty}({\bf x}) \Rightarrow ((a + b) > 1024\cdot c) $$
where the variables are Boolean (the truth value of ${\tt HasProperty}$) and
rational ($a$, $b$, and $c$).  More specifically, SMT is the decision problem
of finding a variable assignment that makes all logical and theory-specific
formulae true, and is analogous to SAT. Recently, researchers have leveraged
SMT for optimization~\cite{cimatti2013modular}. In particular, MAX-SMT requires
to maximize the {\em total weight} of the satisfied formulae; Optimization
Modulo Theories, or OMT, requires to maximize the {\em amount of satisfaction}
of all weighted formulae, and strictly subsumes MAX-SMT.  Most important for
the scope of this paper is that there are high quality MAX-SMT (and OMT)
solvers, which (at least for the $\BV$ and $\LAQ$ theories) can handle problems
with a large number of hybrid variables.

In this paper we propose {\em Learning Modulo Theories} (LMT), a class of novel
hybrid statistical relational learning methods. By combining the flexibility of
structured output Support Vector Machines~\cite{tsochantaridis2005large} and
the expressivity and Satifiability Modulo Theories, LMT is able to perform
learning and inference in mixed Boolean-numerical domains.  Thanks to the
efficiency of the underlying OMT solver, and of the discriminative max-margin
weight learning procedure we propose, we expect LMT to scale to large
constructive learning problems. Furthermore, LMT is {\em generic}, and can in
principle be applied to any of the existing SMT background theories. In the
following two sections we give a short overview of SMT and detail how it can be
employed with the structured output SVM framework, then we describe a few
applications that can be tackled with our approach.

There is relatively little previous work on {\em hybrid} SRL methods. Most current
approaches are direct generalizations of existing SRL
methods~\cite{getoor2007introduction}. Hybrid Markov Logic
networks~\cite{wang2008hybrid} extend Markov Logic by including continuous
variables, and allow to embed numerical comparison operators (namely $\ne$,
$\geq$ and $\leq$) into the constraints by defining an {\em ad hoc} translation
of said operators to a continuous form amenable to numerical optimization.
Inference relies on an MCMC procedure that interleaves calls to a MAX-SAT
solver and to a numerical optimization procedure. This results in an
expensive iterative process, which can hardly scale with the size of the
problem. Conversely, MAX-SMT and OMT are specifically designed to tightly
integrate a theory-specific and a SAT solver, and we expect them to perform
very efficiently. Some probabilistic-logical methods, e.g.
ProbLog~\cite{gutmann2011extending} and PRISM~\cite{islam2012parameter}, have
also been modified to deal with continuous random variables. These models,
however, rely on probabilistic assumptions that make it difficult to implement
fully expressive constraints in, e.g. linear arithmetic, in their formalism.
While there are other interesting hybrid and continuous approaches in the
literature, we skip over them due to space restrictions.

\section{Satisfiability Modulo Theories}
\label{sec:smt}

Propositional satisfiability, or SAT, is the problem of deciding whether a
logical formula over Boolean variables and logical connectives can be satisfied
by some truth value assignment of the variables. Satisfiability Modulo
Theories, or SMT, generalize SAT problems by considering the satisfiability of
a formula with respect to a {\em background theory}
$\mathcal{T}$~\cite{barrett2009satisfiability}. The latter provides the meaning
of predicates and function symbols that would otherwise be difficult to
describe, and reason over, in classical logic. SMT is fundamental in mixed
Boolean domains, which require to reason about equalities, arithmetic
operations and data structures. Popular theories include, e.g. those of linear
arithmetic over the rationals \LAQ\ or integers \LAZ, bit-vectors \BV, strings
\ST, and others.
Most current SMT solvers are based on a very efficient {\em lazy} procedure to
find a satisfying assignment of the Boolean and the theory-specific variables:
the search process alternates calls to an underlying SAT procedure and a
specialized theory-specific solver, until a solution satisfying both solvers is
retrieved, or the problem is found to be unsatisfiable. Recently, researchers
have developed methods to solve the SMT equivalent of MAX-SAT and more complex
optimization problems~\cite{sebastiani2012optimization}. In particular, MAX-SMT
requires to maximize the {\em total weight} of the satisfied formulae;
Optimization Modulo Theories, or OMT, requires to maximize the {\em amount of
satisfaction} of all formulae, modulated by the formulae weights. Clearly, OMT
is strictly more expressive than MAX-SMT.  There are a number of very efficient
MAX-SMT packages available, specialized for a subset of the available theories,
such as MathSAT 5~\cite{cimatti2013mathsat5}, Yices~\cite{dutertre2006yices},
Barcelogic~\cite{bofill2008barcelogic}, which can deal with large problems. SMT
solvers have been previously exploited to perform e.g., formal microcode
verification at Intel~\cite{cimatti2013mathsat5} and large-scale circuit
analysis in synthetic biology~\cite{yordanov2013functional}, and their
optimization counterparts hold much promise. Most important for the goal of
this paper, the MathSAT 5 solver also supports full-fledged OMT problems in the
\LAQ\ theory of linear arithmetic~\cite{sebastiani2012optimization}.

\section{Method Overview}
\label{sec:method}

Structured output SVMs~\cite{tsochantaridis2005large} are a very flexible
framework that generalizes max-margin methods to the case of multi-label
classification with exponentially many classes. In this setting, the
association between inputs ${\bf x}\in\mathcal{X}$ and outputs ${\bf
y}\in\mathcal{Y}$ is controlled by a so-called {\em compatibility function}
$f({\bf x},{\bf y})\equiv{\bf w}^T\bm{\Psi}({\bf x},{\bf y})$,
defined as a linear combination of the joint feature space representation
$\bm{\Psi}$ of the input-output pair and a vector of learned weights ${\bf w}$.
Inference reduces to finding the most compatible output ${\bf
y}^*$ for a given input ${\bf x}$:
\beq
	{\bf y}^* = \argmax_{{\bf y}\in\mathcal{Y}} f({\bf x}, {\bf y}) = \argmax_{{\bf y}\in\mathcal{Y}} {\bf w}^T \bm{\Psi}({\bf x},{\bf y})
\label{eq:inference}
\eeq
Performing inference is non-trivial, since the maximization ranges over an
exponential number of possible outputs.

In order to learn the weights from a training set of $n$ examples
$\left\{({\bf x}_i,{\bf y}_i)\right\}_{i=1}^n\subset\mathcal{X}\times\mathcal{Y}$,
we need to define a non-negative {\em loss function}
$\Delta({\bf y}_i,{\bf y})$
that quantifies the penalty incurred when predicting ${\bf y}$ instead of
the correct output ${\bf y}_i$. Weight learning can then be expressed,
following the {\em margin rescaling}
formulation~\cite{tsochantaridis2005large}, as finding the weights ${\bf w}$
that jointly minimize the training error $\bm{\xi}$ and the model complexity:
\beq
	\argmin_{{\bf w}\in\mathbb{R}^d,\bm{\xi}\in\mathbb{R}^n} \| {\bf w} \|_2 + \frac{C}{n}\sum_{i=1}^n \xi_i
\label{eq:learning}
\eeq
$$	s.t. \quad {\bf w}^T \left( \bm{\Psi}({\bf x}_i,{\bf y}_i) - \bm{\Psi}({\bf x}_i,{\bf y}') \right) \geq \Delta({\bf y}_i,{\bf y}') - \xi_i, \qquad \forall \; i=1,\ldots,n;\;{\bf y}' \neq {\bf y}_i $$
Here the constraints require that the compatibility between ${\bf x}_i$ and the
correct output ${\bf y}_i$ is always higher than that with all wrong outputs
${\bf y}'$, with $\xi_i$ playing the role of per-instance violations.  Weight
learning is a quadratic program, and can be solved very efficiently with a
cutting-plane algorithm~\cite{tsochantaridis2005large}. Since in
Eq~\ref{eq:learning} there is an exponential number of constraints, it is
infeasible to na\"ively account for all of them during learning. Based on the
observations that the constraints obey a subsumption relation, the CP
algorithm~\cite{joachims2009cutting} sidesteps the issue by keeping a working
set of active constraints: at each iteration, it augments the working set with
the most violated constraint, and then solves the corresponding reduced
quadratic program. The procedure is guaranteed to find an
$\epsilon$-approximate solution to the QP in a polynomial number of iterations,
independently of the cardinality of $\mathcal{Y}$ and the number of examples
$n$~\cite{tsochantaridis2005large}.

The CP algorithm is generic, meaning that it can be adapted to any structured
prediction problem as long as it is provided with: i) a joint feature space
representation $\bm{\Psi}$ of input-output pairs (and consequently a compatibility
function $f$); ii) an oracle to perform inference, i.e. Equation~\ref{eq:inference};
iii) an oracle to retrieve the most violated constraint of the QP, i.e. solve
the {\em separation} problem:
\beq
	\argmax_{{\bf y}'} {\bf w}^T\bm{\Psi}({\bf x}_i,{\bf y}') + \Delta({\bf y}_i,{\bf y}')
\label{eq:separation}
\eeq
The oracles are used as sub-routines during the optimization procedure.
Efficient implementations of the oracles are fundamental for the prediction to
be tractable in practice.
For a more detailed exposition, please refer to~\cite{tsochantaridis2005large}.
In the following we provide exactly the three ingredients required to apply the
structured output SVM framework for predicting hybrid boolean-continuous
possible worlds.

We first define the LMT joint feature space of possible words ${\bf z} = ({\bf
x},{\bf y})$. Our definition is grounded on the concept of {\em violation} or
{\em cost} incurred by ${\bf z}$ with respect to a set of SMT formulae. Given
$m$ formulae $F=\{f_j\}_{j=1}^m$, we define the feature vector
$\bm{\Psi}_F({\bf z})\equiv\left( \psi_{f_1}({\bf z}), \;\ldots, \;\psi_{f_m}({\bf z}) \right)$
as the collation of $m$ per-formula cost functions $\psi_f({\bf z})$. In the
simplest case, the individual components $\psi_f$ are indicator functions,
termed {\em boolean costs}, that evaluate to $0$ if ${\bf z}$ satisfies $f$,
and to $1$ otherwise. The LMT compatibility function, written as
$f({\bf z})\equiv{\bf w}^T\bm{\Psi}_F({\bf z})$,
represents the {\em total cost} incurred by a possible world: each
unsatisfied formula $f_j$ contributes an additive factor $w_j$ to
$\bm{\Psi}_F$, while satisfied formulae carry no contribution. Two possible
worlds ${\bf z}$ and ${\bf z}'$ are therefore close in feature space if they
satisfy/violate similar sets of constraints.

Since we want the formulae to {\em hold} in the predicted output, we want to
{\em minimize} the total cost of the unsatisfied rules, or equivalently
maximize its opposite:
$ \argmax_{{\bf y}\in\mathcal{Y}} - {\bf w}^T \bm{\Psi}({\bf x},{\bf y}) $.
The resulting optimization problem is identical to the original inference
problem in Equation~\ref{eq:inference}, as the minus at the RHS can be absorbed
into the learned weights. By defining an appropriate loss function, such as
the Hamming loss $\Delta({\bf y},{\bf y}') \equiv \sum_j \mathbb{I}({\bf y}_j \neq {\bf y}_j')$,
it turns out that both Eq.~\ref{eq:inference} and Eq.\ref{eq:separation} can be
interpreted as MAX-SMT problems. This observation enables us to use a MAX-SMT
solver to implement the two oracles required by the CP algorithm, and thus to
efficiently solve the learning task. Note also that {\em hard} constraints,
i.e. formulae with infinite weight, can also be included in the SMT problem.

The above definition of per-formula {\em boolean cost} $\psi_f$ is only the
simplest option. A more refined alternative, applicable to formulae with only
numerical variables, is to employ a {\em linear cost} of the assignment ${\bf
z}$ and the constants appearing in the formula $f$, as follows:
$$ \psi_{y<c}(y)\equiv\max(y-c,0) \qquad \psi_{y>c}(y)\equiv\max(c-y,0) \qquad \psi_{y=c}\equiv|y-c| $$
For instance, given $f=x+y<5$ and ${\bf z}=(x,y)=(4,3)$, the amount of
violation would be $\psi_f({\bf z})=\max((x+y)-5,0)=2$, while for ${\bf
z}=(1,5)$ the cost would be $0$ (since $f$ is satisfied). Applying linear costs
has two consequences. First, they allow to enrich the feature space with
information about the {\em amount} of violation of any linear formula $f$: an
unsatisfied formula contributes $w_j\cdot\psi_{f_j}({\bf z})$ to $\bm{\Psi}_F$.
Second, since the cost of unsatisfied constraints depends on the value of the
numerical variables involved, the resulting inference and separation oracles
can not be solved using MAX-SMT, but require a full-fledged OMT solver. More
complex cost functions can be developed for mixed boolean-numerical formulae
(consider e.g. $(y > 10) \Rightarrow (a \vee b)$), for instance by summing the
violations of the individual clauses. One issue with this formulation is that,
since the cost of continuous clauses is unbounded, inference may have a bias
towards satisfying them rather than the Boolean ones; this problem however is
shared by all hybrid satisfaction-based models, and its practical impact is not
yet clear.

\section{Applications}
\label{sec:apps}

There are a number of applications involving both Boolean and numerical
constraints, such as environment learning for robot
planning~\cite{wang2008hybrid} and the modeling of gene expression
data~\cite{kuvzelka2011Gaussian}. Here we describe two of them, to illustrate
the flexibility and expressive power of LMT. We postpone a formal definition of
these problems to a future publication, due to space restrictions.

{\em Activity recognition}~\cite{van2008accurate} is the problem of determining
which human activities ${\bf y}_t$ have produced a given set of sensor
observations ${\bf x}_t$ at each time instant $t$. Here the {\em activities}
are understood to be common everyday tasks such as ``having breakfast'',
``watching TV'' or ``taking a shower''. The observations are taken from sensors
deployed in a smart environment (e.g. an instrumented home/hospital), and may
include different sensory channels such as video, audio, the agent's position,
posture, heartbeat, {\em etc.}  Activity recognition is typically cast as a
{\em tagging} problem in discrete time, and tackled by means of probabilistic
temporal models. In real-world scenarios the activities are often concurrent
and inter-related, in which case Factorial versions of Hidden Markov Models or
Conditional Random Fields are used. Unfortunately, training these models is
intractable.  With LMT we take a rather different route, and cast activity
recognition as a form of data-driven {\em scheduling} in continuous time.
Allen's {\em interval temporal logic} (ITL)~\cite{allen1994actions} is an
intuitive formal language to express relations between temporal events. ITL
provides primitives such as {\tt before}$(a,b)$, {\tt after}$(a,b)$, {\tt
overlaps}$(a,b)$, {\tt during}$(a,b)$, {\tt equal}$(a,b)$.  These predicates
can be straightforwardly translated to linear arithmetic constraints, and
therefore easily implemented in LMT. The combination of ITL and FOL allows to
express concurrent, interdependent, nested and hierarchical activities, and to
specify the likely duration of activities and intervals between them. Consider
for instance constraints such as ``breakfast occurs within an hour after waking
up'', and ``cooking a dish involves interacting with at least three
ingredients, in a specific order''.  Using similar constraints, LMT would be
able to generate a scheduling of the activities that is consistent with respect
to the observations and with the (soft) constraints.

Another interesting application is the {\em housing
problem}~\cite{campigotto2011active}, which is just one instance of a class of
weighted constraint satisfaction problems that routinely occur in logistics.
Consider a customer planning to build her own house and judging potential
housing locations provided by a real estate company. There are different
locations available, characterized by different housing values, prices,
constraints about the design of the building (e.g a minimum distance to other
buildings), {\em etc.} A description of the customer preferences and
requirements may be given in SMT, in order to express them with both Boolean
and numerical constraints, e.g., the crime rate, distance from downtown,
location-based taxes, public transit service quality, maxiumum walking or
cycling distances to the closest facilities. The underlying optimization
problem is clearly an instance of MAX-SMT, and LMT can be used to efficiently
learn the formula weights from user-provided data.  We have already developed a
MAX-SMT-based prototype to solve the housing problem in an active learning
setting, by using an interactive preference elicitation mechanism to learn the
relative importance of the various constraints for the customer which has shown
encouraging results~\cite{campigotto2011active}.

\vspace{-4mm}
\bibliographystyle{unsrt}
\bibliography{nips13}

\end{document}